# Multi-scenario highway lane-change intention prediction: A physics-informed AI framework for three-class classification

Jiazhao Shi[*a], Yichen Lin[a], Yiheng Hua[b], Ziyu Wang[c], Zijian Zhang[d], Wenjia Zheng[e], Yun Song[e], Kuan Lu[f], Shoufeng Lu[g]

[a]Tandon School of Engineering, New York University, 6 MetroTech Center, Brooklyn, NY 11201, USA; [b]Cornell Tech, Cornell University, 2 W Loop Rd, New York, NY 10044, USA; [c]School of Business, Wake Forest University, 1834 Wake Forest Rd, Winston-Salem, NC 27109, USA; [d]Department of Computer and Information Science, University of Pennsylvania, Philadelphia, PA 19104, USA; [e]Department of Computer and Information Science, Fordham University, 113 W 60th St, New York, NY 10023, USA; [f]School of Electrical and Computer Engineering, Cornell University, 300 Day Hall, 10 East Avenue, Ithaca, NY 14853, USA; [g]College of Transportation Engineering, Nanjing Tech University, 30 Puzhu South Road, Nanjing, Jiangsu 211816, China

[*]js12624@nyu.edu

## ABSTRACT

Lane-change maneuvers are a leading cause of highway accidents, underscoring the need for accurate intention prediction to improve the safety and decision-making of autonomous driving systems. While prior studies using machine learning and deep learning methods (e.g., SVM, CNN, LSTM, Transformers) have shown promise, most approaches remain limited by binary classification, lack of scenario diversity, and degraded performance under longer prediction horizons.

In this study, we propose a physics-informed AI framework that explicitly integrates vehicle kinematics, interaction feasibility, and traffic-safety metrics (e.g., distance headway, time headway, time-to-collision, closing gap time) into the learning process. lane-change prediction is formulated as a three-class problem that distinguishes left change, right change, and no change, and is evaluated across both straight highway segments (highD) and complex ramp scenarios (exiD). By integrating vehicle kinematics with interaction features, our machine learning models, particularly LightGBM, achieve state-of-the-art accuracy and strong generalization. Results show up to 99.8% accuracy and 93.6% macro F1 on highD, and 96.1% accuracy and 88.7% macro F1 on exiD at a 1-second horizon, outperforming a two-layer stacked LSTM baseline.

These findings demonstrate the practical advantages of a physics-informed and feature-rich machine learning framework for real-time lane-change intention prediction in autonomous driving systems.

## 1. INTRODUCTION

Lane-change maneuvers are a major contributing factor to traffic accidents on highways, with sudden or unsafe lane changes increasing crash risk by 2.53 times [1]. Accurately predicting the lane-change intentions of surrounding vehicles in advance is crucial for improving both the safety and decision-making efficiency of autonomous driving systems [2]. Consequently, real-time lane-change intention prediction has emerged as a highly active research area.

Early studies in this field primarily relied on rule-based approaches and traditional machine learning algorithms, such as Support Vector Machines combined with Hidden Markov Models [3], Gradient Boosted Decision Trees [4], and Bayesian models [5], etc. With the rise of deep learning, research has shifted toward data-driven methods to improve prediction accuracy. Models such as Long Short-Term Memory networks (LSTM) and Convolutional Neural Networks (CNN) have been applied to lane-change intention prediction, achieving notable gains. For instance, Zhang et al. [6] designed a CNN model leveraging multi-channel temporal trajectory data, while Xing et al. [7] proposed an ensemble learning framework based on LSTM that achieved high accuracy. More recently, advanced architecture, such as Transformers, have been employed to further enhance performance. Gao et al. [8] developed a dual-Transformer structure for predicting both lane-change intention and trajectories, while Hu et al. [9] combined residual networks with Transformers for highway merging scenarios—both achieving significant error reductions.

Hybrid approaches have also been explored, integrating multiple models to enhance performance. For example, Geng et al. [10] employed XGBoost to identify lane-changing possibilities and further refined predictions using a bidirectional gated recurrent unit, achieving an overall accuracy of over 98% and outperforming single-model approaches. Wang et al. [11] combined Bi-LSTM with Conditional Random Fields (CRF) and incorporated prior rules to attain high-precision recognition. At the same time, the availability of large-scale, high-resolution trajectory datasets has greatly facilitated model training and evaluation. The highD dataset, collected on German highways, and the exiD dataset, focused on ramp-merging and diverging scenarios, have become widely adopted benchmarks [12][13].

Despite these advancements, several limitations remain:

1. **Binary classification focus**. Most existing work simplifies lane-change intention prediction to a binary classification problem ("change" vs. "no change"). For example, Song and Li [14] categorized all methods into this framework, which fails to capture the dynamic and risk-related distinctions between left and right lane changes.
2. **Limited scenario diversity**. Many models are trained and tested only on straight road segments, with limited attention to complex scenarios such as merging at ramps. Xing et al.'s [7] deep ensemble model performed well in straight-road scenarios but degraded significantly in ramp scenarios; De Cristofaro et al. [15] also noted that vehicle dynamics vary greatly across different highway contexts, and adaptability to complex settings is still insufficient.
3. **Limitation in prediction horizons**. In straight-road dominated prediction scenarios, Many studies show a sharp drop in accuracy as the prediction horizon increases. For instance, Geng et al. [10] reported 98.81% accuracy at 0 s horizon but only 87% at 2 s. Zhang and Fu [16] achieved 93% at a 3 s horizon via online transfer learning, but their model remained binary and could not distinguish left from right lane changes.

Motivated by these gaps, this study addresses multi-class lane-change intention prediction (left, right, no change) across multiple highway scenarios with extended prediction horizons. The main contributions are:

1. **Physics-Informed AI framework.** We designed a physics-guided feature pipeline grounded in vehicle kinematics and traffic-flow safety theory (e.g., distance headway, time headway, time-to-collision, closing gap time), neighbor-interaction safe gaps, lane-advantage indices, and the proposed CGT measure. This yields stronger generalization and more stable minority-class detection across horizons. In cross-location evaluations (The model was evaluated on data from specific locations: from the highD dataset, locations 4 and 5 were used; from the exiD dataset, locations 4, 5, and 6 were used.), LightGBM model, as the best model, achieves overall accuracies of approximately 99.8%, 99.5%, and 99.1% at 1 s, 2 s, and 3 s horizons on highD, and 96.1%, 89.5%, and 81.2% on exiD.
2. **Multi-class classification**. By distinguishing between left lane change, right lane change, and no lane change, the models can more precisely capture directional dynamics. They achieve macro F1-scores of 93.6%, 91.3% and 87.6% for highD at 1 s, 2 s, and 3 s horizons, and 88.7%, 83.4%, and 78.4% for exiD, respectively.
3. **Scenario-specific analysis**. We evaluate performance on both straight segments (highD) and ramp merging scenarios (exiD), revealing differences in prediction difficulty and feature importance. Our model outperforms comparable studies and demonstrates strong cross-scenario generalization, providing a robust foundation for real-time decision-making in intelligent connected vehicles.
4. **Comparative evaluation with deep learning**. We benchmark LightGBM and XGBoost against a two-layer stacked LSTM. Results show that, under current data conditions, traditional machine learning models offer superior prediction accuracy, training efficiency, and generalization. This finding further demonstrates the practical applicability of structurally simple and feature-interpretable machine learning methods for lane-changing intention prediction.

Overall, our work achieves state-of-the-art accuracy while delivering strong cross-scenario robustness, providing a physics-informed and practically deployable solution for highway autonomous driving systems.

## 2. METHODS

Unlike most existing studies, we formulate lane-change prediction as a three-class classification problem, where "left lane change," "right lane change," and "no lane change" are treated as distinct categories. Using the highD and exiD datasets, our goal is to predict whether a given vehicle will execute any of these lane-change maneuvers within a specified prediction horizon.

### 2.1. Physics-informed AI framework

This study proposes a Physics-Informed AI framework shown in Figure 1 that systematically embeds domain knowledge from vehicle kinematics, traffic safety, and roadway dynamics into the learning pipeline. As a primary innovation of this work, this framework serves as the foundational architecture, within which specific machine learning and deep learning models (detailed in subsequent sections) are implemented and evaluated. Unlike purely data-driven approaches, our framework explicitly incorporates physical plausibility and interaction feasibility as inductive biases, thereby enhancing model generalization, stability, and interpretability across diverse highway scenarios.

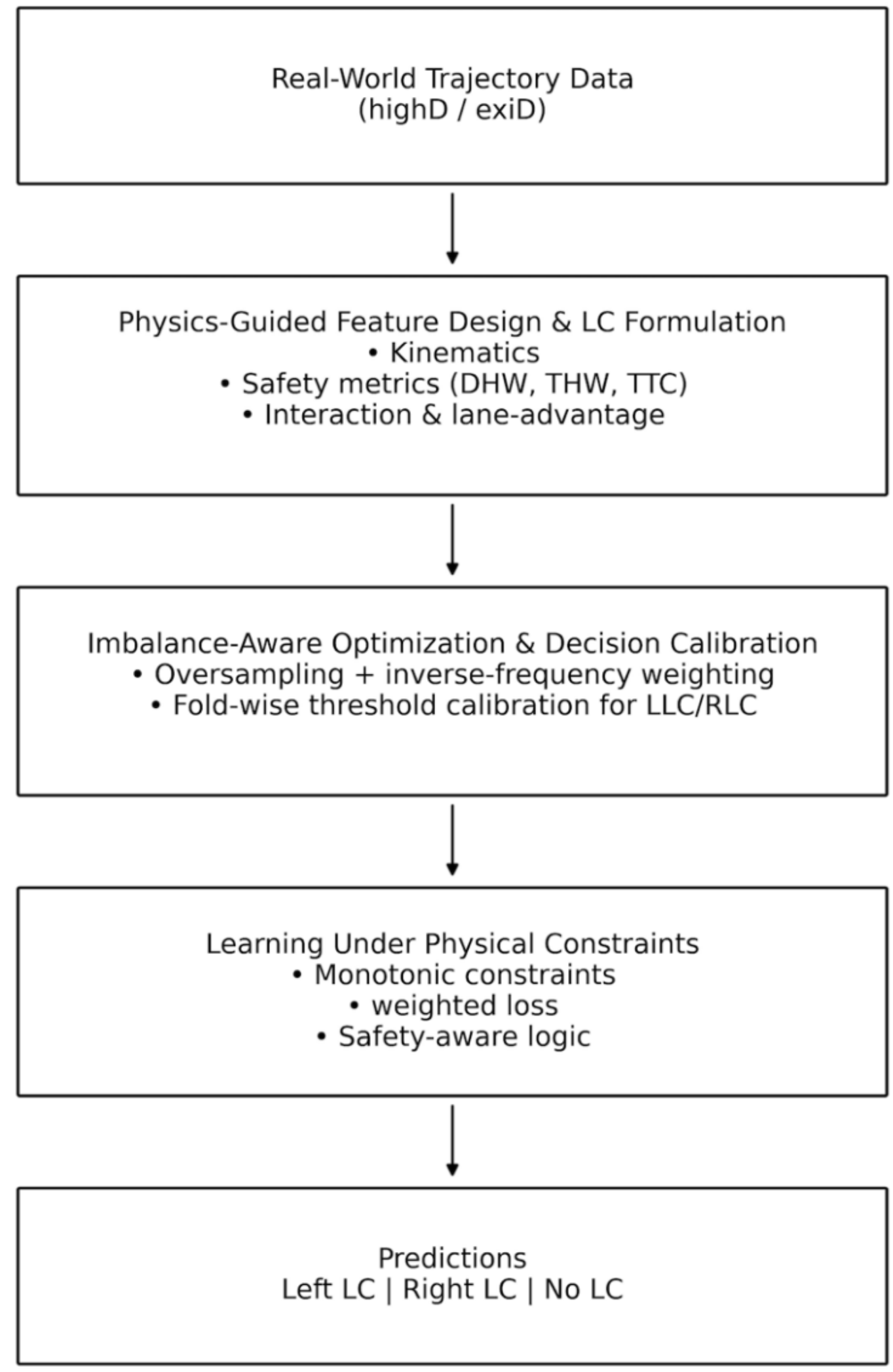

Figure 1. Structural diagram of the proposed physics-informed AI framework.

The framework consists of three interconnected components, which guide the entire process from problem formulation to model validation:

1. **Physics-Guided Feature Design and Lane-Change Formulation.** The input space is extended beyond raw trajectory data through the inclusion of features derived from domain knowledge. This includes vehicle kinematics (e.g., speed, acceleration, yaw rate), longitudinal safety metrics such as Distance Headway (DHW), Time Headway (THW), and Time-to-Collision (TTC), as well as interaction-aware indicators like safe gap availability and lane-advantage indices.

These features collectively capture the dynamic context and risk landscape surrounding the ego vehicle, providing a physically meaningful basis for maneuver prediction. The detailed definitions and computations of these features are elaborated in Section 3.3.

Within this physics-informed context, we formulate lane-change prediction as a three-class classification problem, distinguishing between "left lane change," "right lane change," and "no lane change." The definitions for these maneuvers are established as follows, tailored to different highway scenarios using the highD and exiD datasets:

**a. Straight Road Scenarios**

Figure 2 shows a lane-changing scenarios on a straight road.

Start Time: The vehicle's lateral coordinate crosses the current lane center line by at least 0.2 m and maintains a monotonic lateral drift for at least 0.5 seconds thereafter.

End Time: The vehicle has fully entered the adjacent lane, with no reverse lateral movement for at least 1 second afterward.

Direction: Under the condition that the driving direction remains consistent, a lane change is considered to the left if the laneId before the change is less than that after the change, and to the right otherwise. (In highD, lanes are numbered from the leftmost lane in each direction, increasing toward the right.).

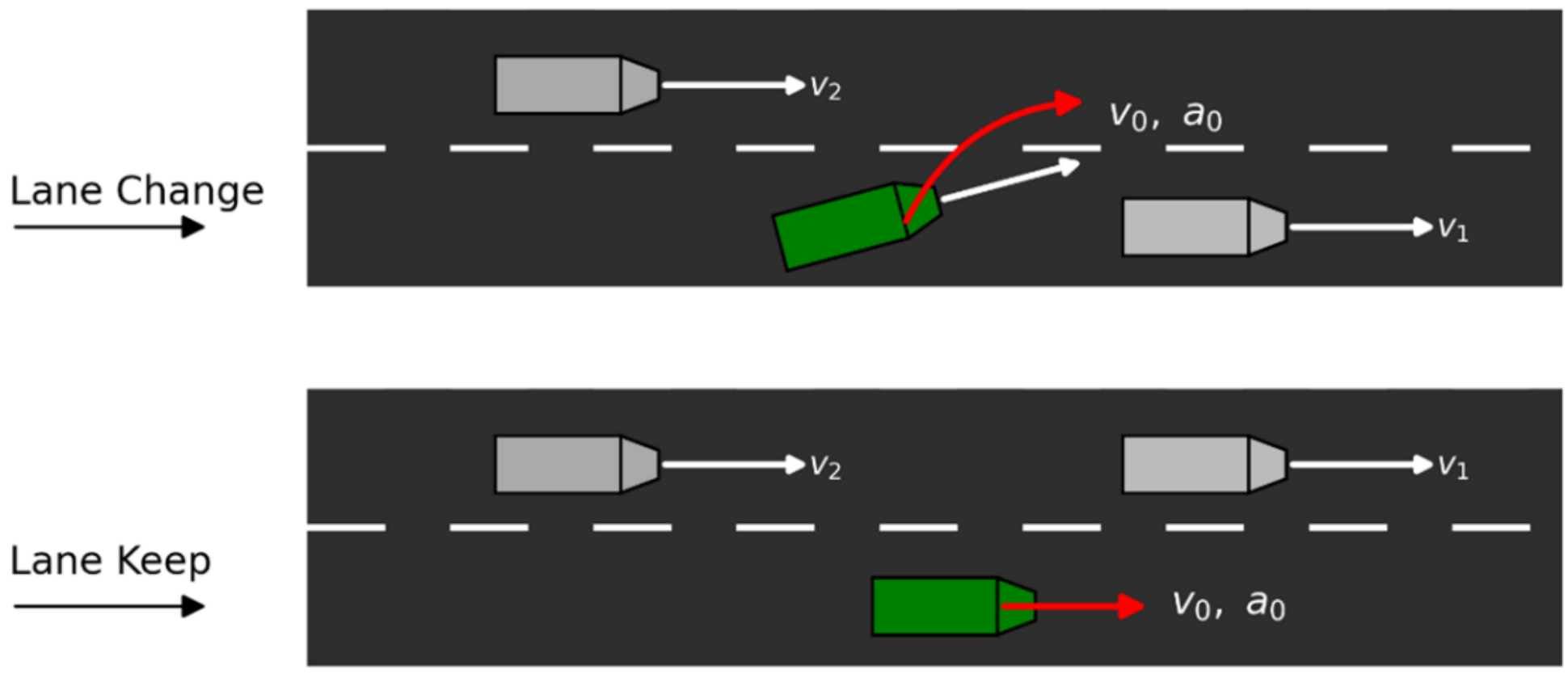

Figure 2. Straight road scenarios (based on highD dataset).

**b. Ramp Merging/Diverging Scenarios**

Figure 3 shows a lane-changing scenarios on a ramp merging/diverging area.

Start Time: The vehicle's lateral coordinate crosses the lane center line by at least 0.2 meters and maintains a monotonic lateral drift for at least 0.5 seconds thereafter.

End Time: The vehicle has fully entered the adjacent lane, with no reverse lateral movement for at least 1 second afterward.

Direction: Due to the non-sequential nature of laneId between the mainline and ramps in the exiD dataset, the sign of the lateral velocity (latVelocity) is used as the criterion instead. The direction is determined by averaging the lateral velocity over a 0.1-second window starting from the start time. If the average lateral velocity is greater than zero, the maneuver is considered a left lane change; otherwise, it is considered a right lane change.

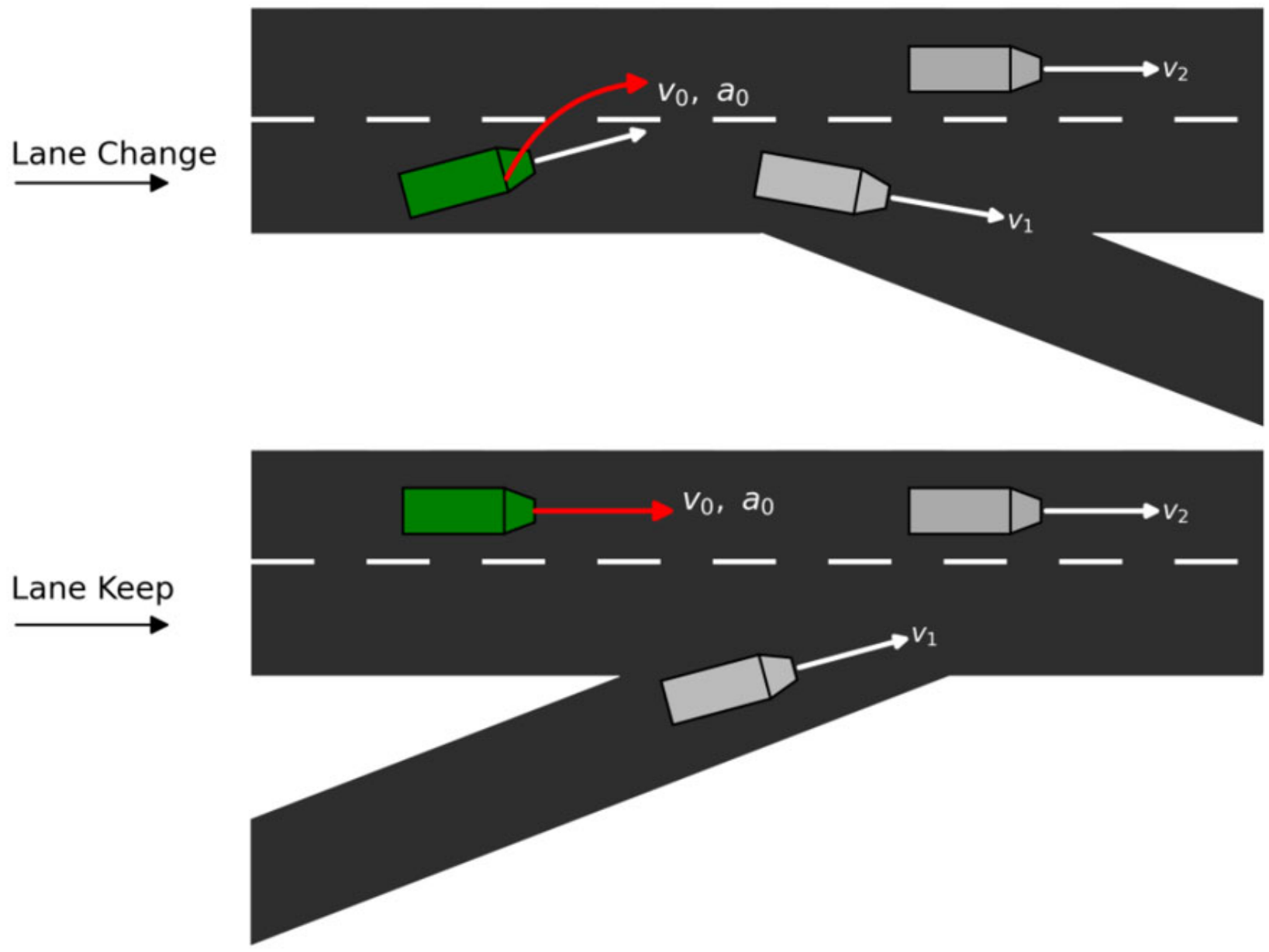

Figure 3. Ramp merging/diverging scenarios (based on exiD dataset).

2. **Imbalance-Aware Optimization and Decision Calibration.** To address class skew and operating-point sensitivity, we combine oversampling and inverse-frequency weighting with fold-wise threshold calibration for minority classes (LLC/RLC) strictly within cross-validation, which improves macro-F1, and minority-class recall without leakage.

3. **Learning Under Physical Constraints.** During model training, we enforce monotonic relationships and safety-aware logic to align predictions with traffic dynamics. For instance, the probability of a lane change is constrained to increase with the number of available safe gaps or with improved lane advantage indices, and to correlate positively with lateral movement toward a target lane. In tree-based models such as LightGBM and XGBoost, these are implemented as monotonicity constraints on specific features, while for the LSTM, they are reflected through feature engineering and weighted loss functions that prioritize safety-critical scenarios.

By unifying problem formulation, feature design, imbalance-aware decision calibration, and constrained learning under a physics-informed paradigm, the proposed framework bridges the gap between data-driven prediction and domain knowledge. The subsequent sections (2.2–2.4) detail the specific algorithmic modules (XGBoost, LightGBM, and a two-layer Stacked LSTM) that are instantiated and evaluated within this overarching framework, resulting in more robust and interpretable lane-change intention recognition suitable for real-world autonomous driving systems.

### 2.2. XGBoost

XGBoost (Extreme Gradient Boosting) is a tree-based machine learning algorithm that incorporates both second-order gradient information and explicit structural regularization. Its design aims to maximize predictive accuracy while preventing overfitting and improving training efficiency [17]. The model optimizes the structure and leaf weights of each new tree directly in the objective space by leveraging both first- and second-order derivatives. For a set of regression trees $f_k \in F$, the overall model prediction is $\hat{y}_i = \phi(x_i) = \sum_{k=1}^{K} f_k(x_i)$, and the regularized learning objective is:

$$L(\varphi) = \sum_{i} \ell(\hat{y}_i, y_i) + \sum_{k} \Omega(f_k), \qquad \Omega(f) = \gamma T + \frac{1}{2}\lambda \|w\|^2 \tag{1}$$

Here, $\ell(\cdot)$ denotes the training loss, $\Omega(f)$ penalizes both the number of leaves $T$ and the squared $L_2$ norm of the leaf weights $w$, and $\gamma, \lambda$ are regularization coefficients.

At each boosting iteration, XGBoost applies a second-order Taylor expansion to approximate the objective as a weighted quadratic function of the first-order gradients $g_i$ and second-order gradients $h_i$. This transforms the split search problem into a regularized optimization problem. By incorporating both “loss reduction” and “curvature” terms, XGBoost improves the reliability of the chosen splits. For a candidate splits that partitions a parent node into left (L) and right (R) child nodes, the loss reduction after the split is given by:

$$Gain = \frac{1}{2}\left(\frac{G_L^2}{H_L+\lambda}+\frac{G_R^2}{H_R+\lambda}-\frac{(G_L+G_R)^2}{H_L+H_R+\lambda}\right)-\gamma \tag{2}$$

where $G = \sum g_i$ and $H = \sum h_i$, are the aggregated gradient statistics for a node. This formula directly quantifies the improvement in node purity obtained by the split while accounting for model complexity via the $-\gamma$ term.

From an engineering perspective, XGBoost incorporates several optimizations, including sparsity-aware split finding (with automatic handling of missing values), block-based data storage for cache efficiency, histogram-based split enumeration, shrinkage (learning rate), and column subsampling for additional regularization. The implementation also supports distributed computing for large-scale datasets and can be applied to regression, binary classification, multi-class classification (via Softmax), and ranking tasks.

In summary, XGBoost's combination of second-order optimization, structural regularization, and computational efficiency allows it to achieve high predictive accuracy with strong generalization. This makes it suitable for industrial-scale machine learning applications where both performance and interpretability are important.

### 2.3. LightGBM

LightGBM (Light Gradient Boosting Machine) is a gradient boosting framework that grows trees in a **leaf-wise** manner, guided by histogram-based feature binning. The core idea is to first discretize continuous features into bins and then perform split searches on the binned data, which significantly reduces computation and memory costs [18]. For any given feature bin $b$, LightGBM aggregates first- and second-order gradient statistics at the node level as follows:

$$G_{k,v} = \sum_{j\in\{j|s_{k,v}\geq x_{jk}>s_{k,v-1}\}} g_j\,, \qquad H_{k,v} = \sum_{j\in\{j|s_{k,v}\geq x_{jk}>s_{k,v-1}\}} h_j \tag{3}$$

Here, $G_{k,v}$ and $H_{k,v}$ represent the cumulative gradients and Hessians for bin v of feature k. This histogram-based aggregation allows for rapid evaluation of split gains, where the gain formula remains consistent with that in GBDT:

$$Gain = \frac{1}{2}\left(\frac{G_L^2}{H_L+\lambda}+\frac{G_R^2}{H_R+\lambda}-\frac{(G_L+G_R)^2}{H_L+H_R+\lambda}\right)-\gamma \tag{4}$$

With $G = \sum g_i$ and $H = \sum h_i$ for each node.

LightGBM's leaf-wise growth strategy expands the leaf with the highest potential gain at each step, typically achieving larger loss reduction with fewer levels compared to depth-wise growth. The framework enforces constraints such as maximum depth and minimum leaf data to prevent overfitting, while ensuring efficiency for large-scale tasks. Furthermore, LightGBM introduces techniques such as Gradient-based One-Side Sampling (GOSS), which retains instances with larger gradients while randomly sampling from small-gradient instances, and Exclusive Feature Bundling (EFB), which combines mutually exclusive features to reduce dimensionality.

For multi-class classification, LightGBM minimizes the following objective:

$$L_{multi} = -\frac{1}{n}\sum_{i=1}^{n}\sum_{c=1}^{C} y_{ic} log p_{ic}\,,\;\; p_{ic} = \frac{\exp\,(z_{ic})}{\sum_{c'} \exp\,(z_{ic'})} \tag{5}$$

Here, $y_{ic}$ is the indicator for class c, and $p_{ic}$ is the predicted probability obtained via the Softmax function.

With its combination of histogram-based binning, leaf-wise growth, and efficiency-oriented optimizations, LightGBM delivers high accuracy and training speed, making it well-suited for large-scale and latency-sensitive applications.

### 2.4. Two-layer stacked LSTM

Long Short-Term Memory (LSTM) is a special type of recurrent neural network (RNN) architecture. Its core idea is to mitigate the vanishing or exploding gradient problems that occur during the training of traditional RNNs by introducing a sophisticated gating mechanism and a cell state, thereby effectively capturing long-term dependencies in time series data [19]. Proposed by Hochreiter and Schmidhuber in 1997, this model has become one of the most influential architectures for sequence modeling tasks.

Let the input sequence be $x = (x_1, x_2, \dots, x_T)$. At each time step $t$, the LSTM unit contains a cell state $C_t$ and a hidden state $h_t$. Its key innovation lies in three gating structures: the forget gate $f_t$, the input gate $i_t$, and the output gate $o_t$. Their mechanisms are described as follows:

**Forget gate** controls the retention degree of the cell state from the previous time step:

$$f_\mathrm{t} = \sigma(\mathrm{W}_f \mathrm{x}_t + \mathrm{U}_f h_{t-1} + \mathrm{b}_f) \tag{6}$$

**Input gate** controls the degree to which the candidate value $\tilde{C}_t$ updates the cell state:

$$i_t = \sigma(\mathrm{W}_i \mathrm{x}_t + \mathrm{U}_i h_{t-1} + \mathrm{b}_i) \tag{7}$$

$$\tilde{C}_\mathrm{t} = \tanh\,(\mathrm{W_c} \mathrm{x}_t + \mathrm{U_c} h_{t-1} + \mathrm{b_c}) \tag{8}$$

**Cell state** undergoes self-recursive update based on the forget gate and the input gate:

$$C_t = f_t \odot c_{t-1} + i_t \odot \tilde{c}_t \tag{9}$$

**Output gate** controls the contribution of the cell state to the current hidden state:

$$\mathrm{o}_t = \sigma(\mathrm{W_o} \mathrm{x}_t + \mathrm{U_o} h_{t-1} + \mathrm{b_o}) \tag{10}$$

$$h_t = \mathrm{o}_t \odot \tanh\,(\mathrm{c}_t) \tag{11}$$

Here, $\sigma(\cdot)$ denotes the sigmoid activation function, $\odot$ represents element-wise multiplication, and $W$ and $b$ are the corresponding weight matrices and bias terms.

Through the above gating units, LSTM achieves explicit regulation of information flow: the forget gate determines the proportion of historical information to discard, the input gate filters new information to incorporate into the cell state, and the output gate controls the degree to which the internal state is exposed externally. This mechanism enables LSTM to selectively maintain or update its memory content, thereby overcoming the challenge of learning long-term dependencies.

At the model architecture level, LSTM can be stacked to form deep networks and supports bidirectional extension (Bi-LSTM) to capture both forward and backward contextual information. At the training level, its parameters can be efficiently optimized through the Backpropagation Through Time (BPTT) algorithm. LSTM is widely applicable to various sequence modeling tasks such as machine translation, speech recognition, time series prediction, and text generation. It demonstrates significant advantages especially in handling long sequences and complex context modeling.

By virtue of its gated recurrent unit structure and the gradient-stabilizing characteristics of the cell state, LSTM effectively addresses the long-term memory problem of traditional RNNs, providing a powerful and general computational framework for modeling sequential data [20].

In this study, a two-layer stacked LSTM architecture is adopted to process frame-level feature sequences, with the following configuration (Shown in Figure 4):

1. Temporal expansion – For a sequence length $T = f_s \times W_h$ (sampling frequency $f_s$ =25 Hz, historical window $W_h$ = 5 s or 6 s), a segment of 90–104 frames is extracted and z-score normalized across 90–104 feature dimensions.

2. Encoding – The first LSTM layer (256 units) reads the input sequence and produces hidden states $h_{1:T}^{(1)}$; the second LSTM layer (128 units) continues encoding these states sequentially.

3. Attention pooling – The output sequence $h_{1:T}^{(2)}$ is processed with an attention mechanism to produce a fixed-dimensional weighted representation, enhancing sensitivity to key frames.

4. Classification – The final representation is passed through a fully connected layer with a SoftMax activation to output probabilities for the three classes *($p_{NLC}$, $p_{LLC}$, $p_{RLC}$)*.

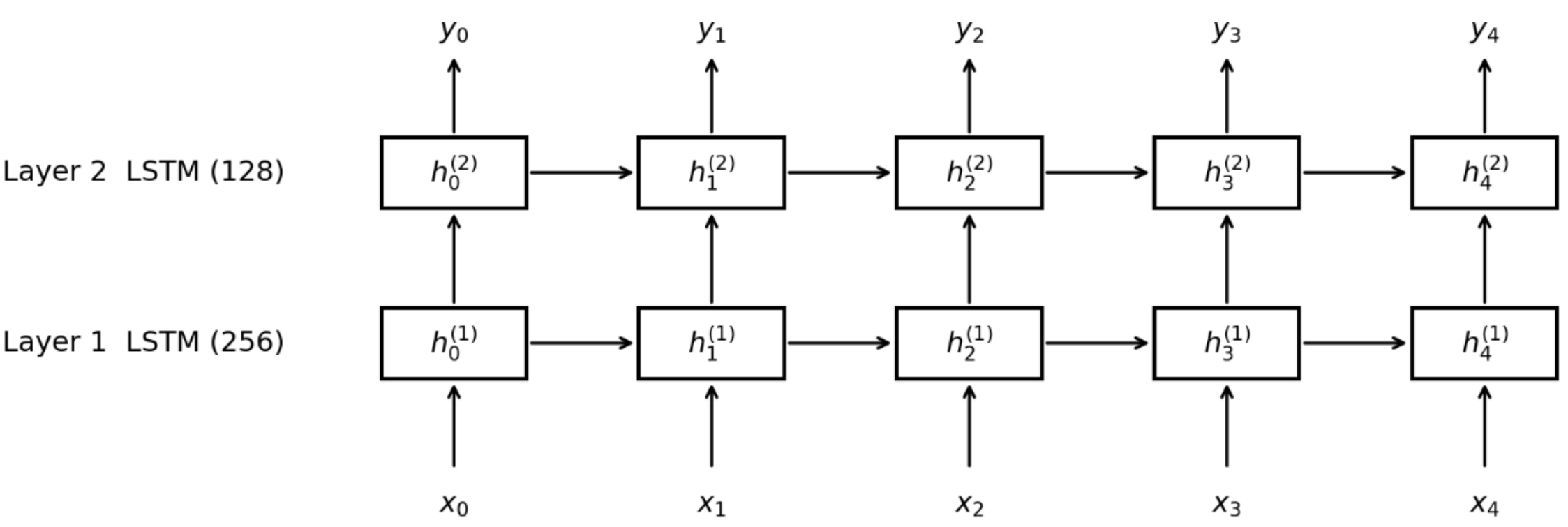

Figure 4. Two-layer stacked LSTM (unrolled over time) – sequence-to-sequence.

## 3. DATA PROCESSING AND FEATURE EXTRACTION

### 3.1. Overview of the highD and exiD datasets

**highD Dataset**. Collected by the ika research team at RWTH Aachen University, the highD dataset consists of aerial recordings from six straight highway segments without ramps. The footage, captured using drones, comprises 60 video sequences (≈16.5 hours, covering approximately 110,500 vehicles). The top-down perspective eliminates occlusion and provides sub-decimeter (<0.1 m) localization accuracy, along with precise road geometry information, making it particularly well-suited for studying lane-changing behavior on conventional straight highway segments.

Dataset link: https://levelxdata.com/highD-dataset/

**exiD Dataset**. The exiD dataset follows the same data collection and annotation procedures as highD but focuses specifically on highway ramp merging and diverging zones, each approximately 420 meters in length. Recorded at seven different highway interchanges in Germany, the dataset comprises 92 aerial video sequences (≈16 hours, covering about 69,172 vehicles). The recordings are processed at 25 Hz and include additional annotations for ramp types and more complex lane topology information, making it particularly valuable for studying traffic behavior in intricate merging and diverging scenarios.

Dataset link: https://levelxdata.com/exiD-dataset/

To mitigate the influence of "bidirectional lane changes" on the decision boundary, we excluded trajectories that exhibited both left and right lane changes within the prediction horizon during the data cleaning stage. Only trajectories with a clear single-direction lane change or no lane change were retained. This filtering step ensures consistent label definitions and reduces class ambiguity caused by mixed-behavior samples.

### 3.2. Data imbalance

In supervised learning, when the distribution of training samples is highly imbalanced, the model tends to minimize the overall loss by favoring the majority classes. While this can lead to high overall accuracy, it often reduces the model's ability to correctly classify minority classes.

This challenge is particularly evident in the highD dataset. Depending on the observation horizon, the imbalance varies considerably: under a short 1-second horizon, the ratio of "no lane change" to "left/right lane change" samples can be as extreme as 250:1:1, whereas extending the horizon to 3 seconds increases the number of lane change instances and reduces the ratio to about 80:1:1. Correspondingly, left and right lane changes each account for only about 0.4% to 1.2% of the samples.

By contrast, in the exiD dataset, the class distribution is notably more balanced. The proportions of "no lane change," "left lane change," and "right lane change" range approximately from 8:1:1 to 2:1:1. Given this relatively balanced distribution, exiD does not require the same degree of imbalance treatment, and we excluded data imbalance handling steps for that dataset.

Therefore, in our preprocessing pipeline, we placed particular emphasis on addressing class imbalance for highD before model training. Our strategy combined three complementary techniques—oversampling, class weighting, and threshold calibration—applied in a layered manner:

(1) Oversampling expands the minority class while preserving the original distribution structure. We use the SMOTE-Tomek method, which generates new samples by randomly interpolating along the lines between minority class instances and their k nearest neighbors. This effectively mitigates the overfitting risk associated with simple replication, resulting in a final sample ratio of approximately 29:1:1 for non-lane-changing, left lane-changing, and right lane-changing instances in the highD dataset.

(2) Class weighting – We introduced inverse frequency weights into the loss function:

$$w_i = (\frac{N}{K\, n_i})^{\alpha}, \alpha \in [0,1] \tag{12}$$

where $N$ is the total number of training samples, $K$ is, the number of classes, and $n_i$ is the number of samples in class $i$. Setting $\alpha = 0.5$ provided a good balance between smoothing and penalization, helping to suppress bias toward the majority classes while improving recall for the minority classes.

(3) Threshold Calibration: For classifiers with probabilistic outputs, the decision threshold is independently optimized on each validation fold during cross-validation. Particularly in cost-sensitive scenarios, the detection threshold for minority classes (LLC and RLC) is appropriately lowered to improve recall without excessively sacrificing overall accuracy. This process is strictly embedded within cross-validation to prevent information leakage.

After the oversampling approach, supplemented with inverse frequency weighting and threshold adjustment, the class distribution in the training set is significantly balanced. The macro F1-score on unseen data increased by an average of 25 percentage points, and the recall rates for the minority classes improved by an average of 40% (LLC) and 32% (RLC). Overall, this stratified processing strategy effectively suppresses overfitting while markedly enhancing the detection capability for minority classes, establishing a solid data foundation for real-time lane-changing intention prediction models.

### 3.3. Feature extraction

This study performs feature engineering using the large-scale, real-world trajectory datasets highD and exiD, both of which provide rich and structured information on vehicle kinematics and roadway geometry. To meet the requirements of real-time lane-change intention prediction, we first applied a unified cleaning procedure to the raw data, synchronized all time references, and expanded the feature set through sequence-based statistics and spatial semantics. The resulting high-dimensional feature vectors comprise 99 features for highD (27 raw variables + 72 derived features) and 104 features for exiD (28 raw variables + 76 derived features). For consistency, reproducibility, and scalability, all features are organized into five major categories, which serve as the physics-guided variables referenced in Section 2.1:

1. Kinematics and temporal statistics. Based on the trajectory of the vehicle's center point, we calculate scalar speed, acceleration magnitude, heading angle, yaw rate, and curvature radius. In addition, we extract short-term dynamics by computing the mean, standard deviation, and extrema of speed, acceleration, and yaw angle within a 1-second rolling window, capturing transient driving trends and stability.

2. Lane position semantics. Using lane IDs, boundary coordinates, and centerline geometry, we derive lateral offsets, distances to lane boundaries, and their first-order differences. We also compute sliding-window means and absolute displacements to characterize lateral confinement and the preparatory phase before a lane change.

3. Neighbor vehicle interactions. Leveraging the neighbor indices provided in the datasets, we construct a local traffic topology and calculate relative longitudinal/lateral distances, speed differences, acceleration differences, and approach rates for six key positions (front/rear/left/right). We further compute the occupancy ratio of neighboring vehicles to quantify local traffic density.

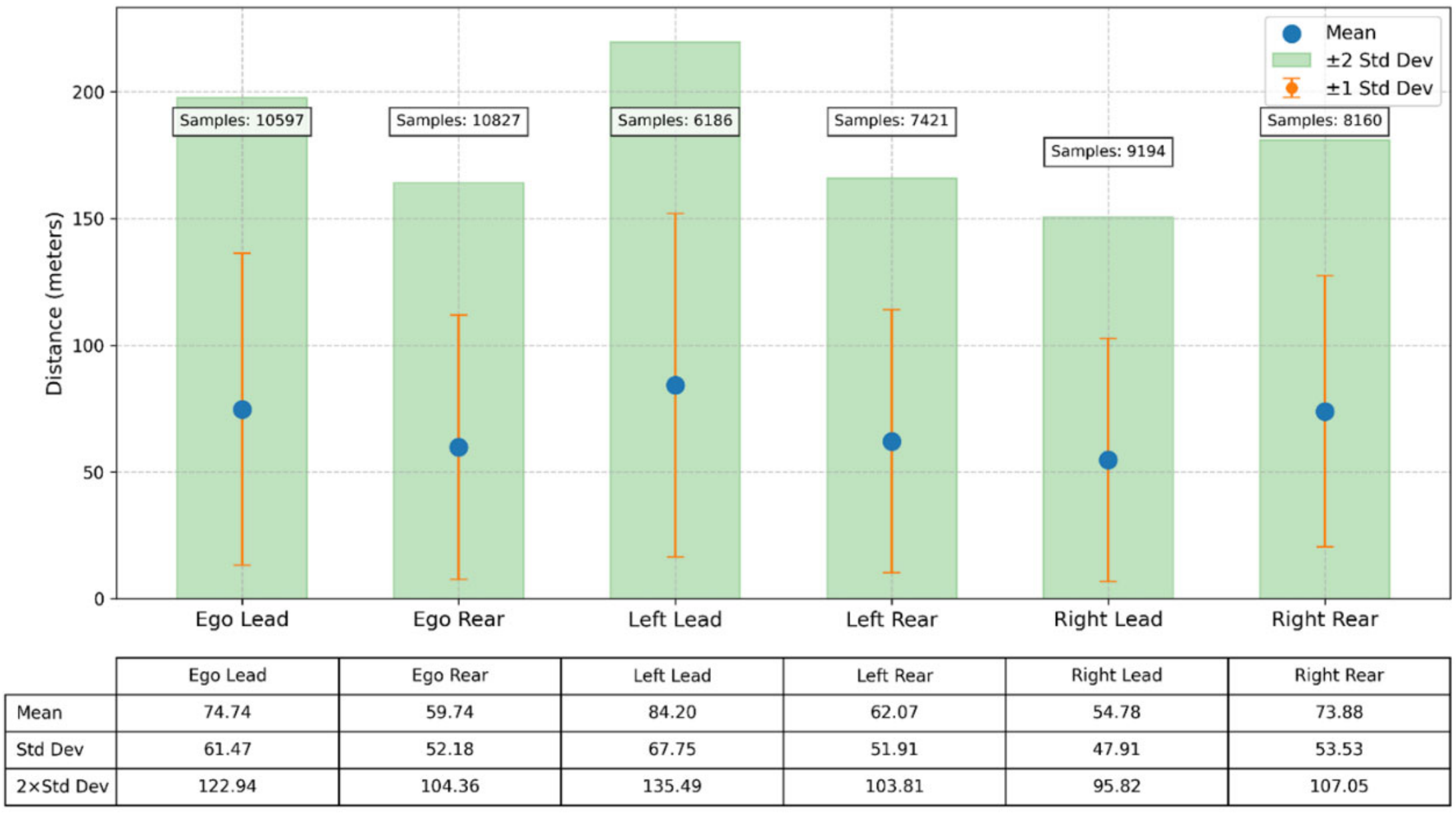

| | Ego Lead | Ego Rear | Left Lead | Left Rear | Right Lead | Right Rear |
|---|---|---|---|---|---|---|
| Mean | 74.74 | 59.74 | 84.20 | 62.07 | 54.78 | 73.88 |
| Std Dev | 61.47 | 52.18 | 67.75 | 51.91 | 47.91 | 53.53 |
| 2×Std Dev | 122.94 | 104.36 | 135.49 | 103.81 | 95.82 | 107.05 |

Figure 5. Average distance and distribution statistics between the lane-changing vehicle and adjacent vehicles (based on the highD dataset).

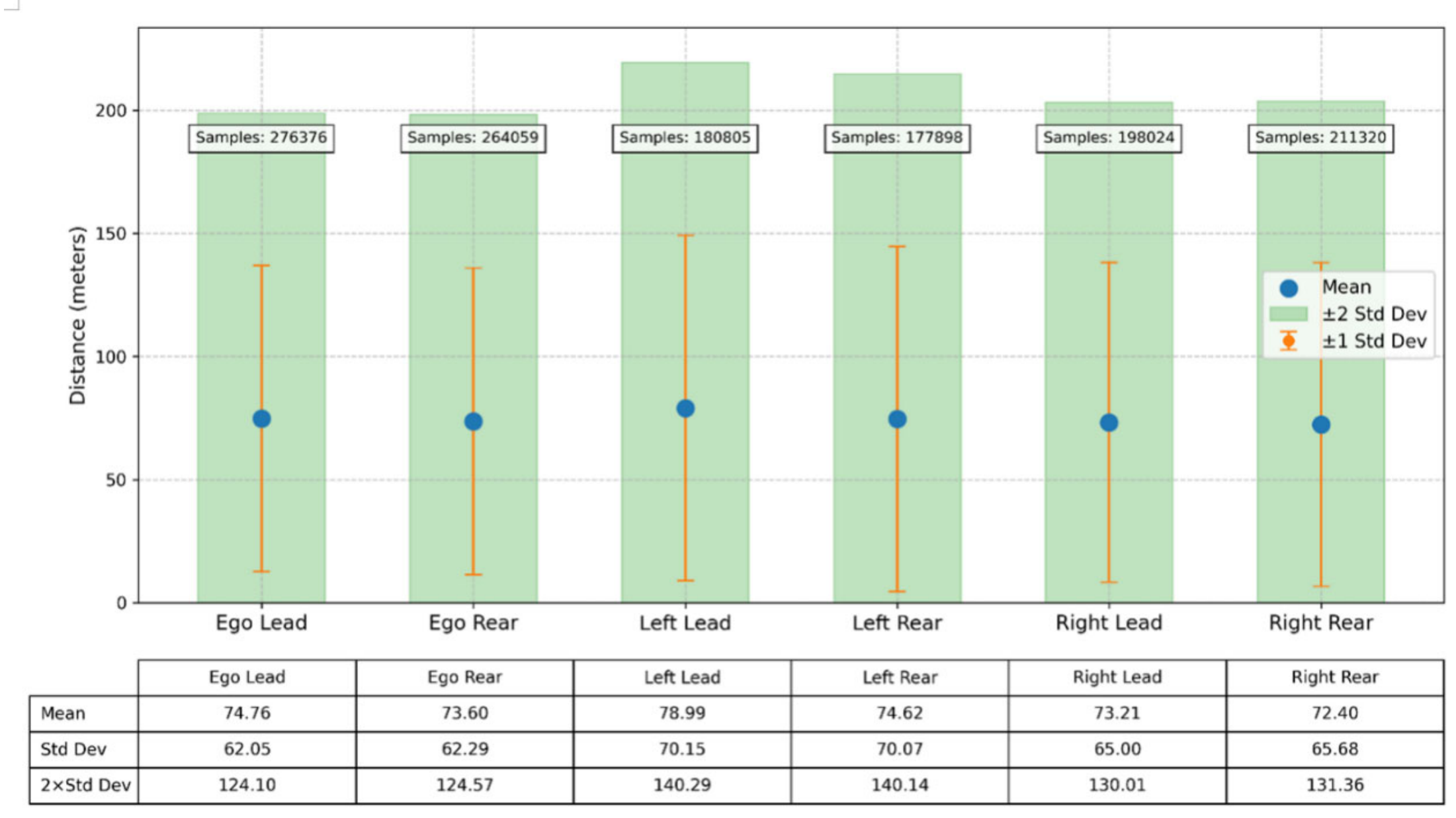

| | Ego Lead | Ego Rear | Left Lead | Left Rear | Right Lead | Right Rear |
|---|---|---|---|---|---|---|
| Mean | 74.76 | 73.60 | 78.99 | 74.62 | 73.21 | 72.40 |
| Std Dev | 62.05 | 62.29 | 70.15 | 70.07 | 65.00 | 65.68 |
| 2×Std Dev | 124.10 | 124.57 | 140.29 | 140.14 | 130.01 | 131.36 |

Figure 6. Average distance and distribution statistics between the lane-changing vehicle and adjacent vehicles (based on the exiD dataset).

Figures 5 and 6 illustrate the lateral distance distributions between the ego vehicle and its surrounding neighbors (front, rear, and adjacent lanes) in the highD and exiD datasets at the moment of lane change. For example, in Figure 5 the average spacing ranges between 54.8–84.2m, with a standard deviation of 47.9–67.8m. The light green region corresponds to $\pm 2\ \sigma$, while the orange bar denotes $\pm 1\ \sigma$, providing empirical thresholds for acceptable safety gaps. Beyond the previously defined interaction features, this distributional information combined with the neighbor IDs in the datasets allows us to extract the following additional indicators:

(a) Normalized distance distribution. For each distance $d_i$, we computed:

$$z_i = \frac{d_i - \mu_i}{\sigma_i}, s_i = \frac{d_i}{\mu_i} \tag{13}$$

Where $\boldsymbol{\mu_i}$ and $\boldsymbol{\sigma_i}$ are the mean and standard deviation of Figure 5 and Figure 6, respectively. The metric $z_i$ reduces scale bias due to roadway or traffic flow differences, while $s_i$ preserves relative spacing information.

(b) Safe gap indicator. A distance is considered safe if $\mathrm{d_i} > \mu_\mathrm{i} + 2\sigma_\mathrm{i}$ (beyond the upper bound of the green region).

$$g_i = \begin{cases} 1, & if\ safe \\ 0, & otherwise \end{cases}, \sum_i g_i\ gives\ the\ number\ of\ safe\ gap. \tag{14}$$

(c) Lane advantage index:

$$\Delta_{\mathrm{lead}} = d_{\mathrm{LeftLead}} - d_{\mathrm{EgoLead}}, \Delta_{\mathrm{rear}} = d_{\mathrm{LeftRear}} - d_{\mathrm{EgoRear}} \tag{15}$$

We define $\min(\Delta_{\mathrm{lead}}, \Delta_{\mathrm{rear}})$ as the overall availability score for the adjacent lane. A positive value indicates that the neighboring lane provides larger longitudinal gaps, implying higher feasibility for a lane change.

(d) Time-to-gap measure. The spacing $d_i$ is divided by the ego vehicle's velocity, producing the residual following time $t_i = d_i / v_{\mathrm{ego}}$. Applying z-score normalization to $t_i$ provides a velocity-independent temporal measure of gap availability.

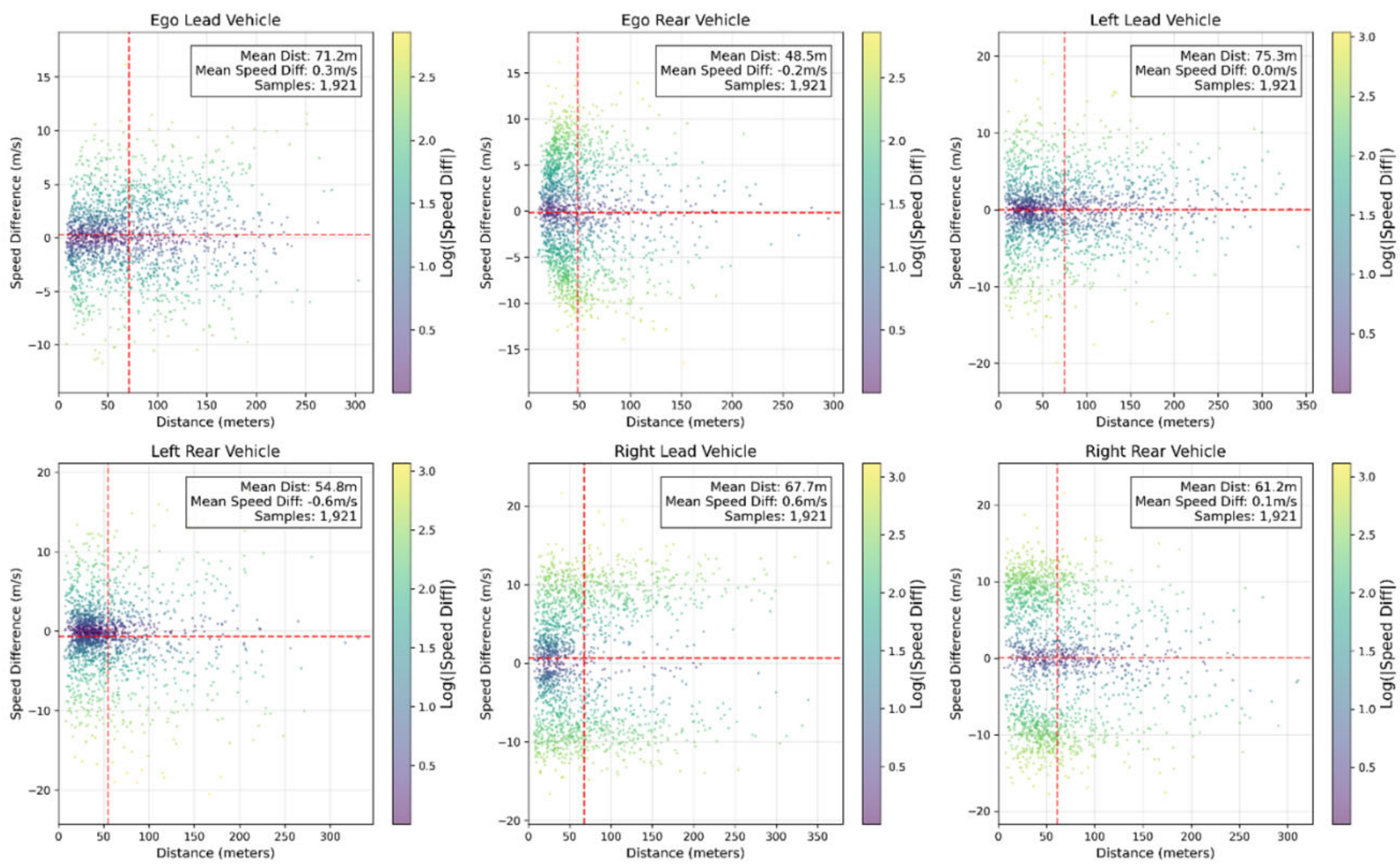

Figure 7. Distance-speed difference distribution (based on the highD dataset).

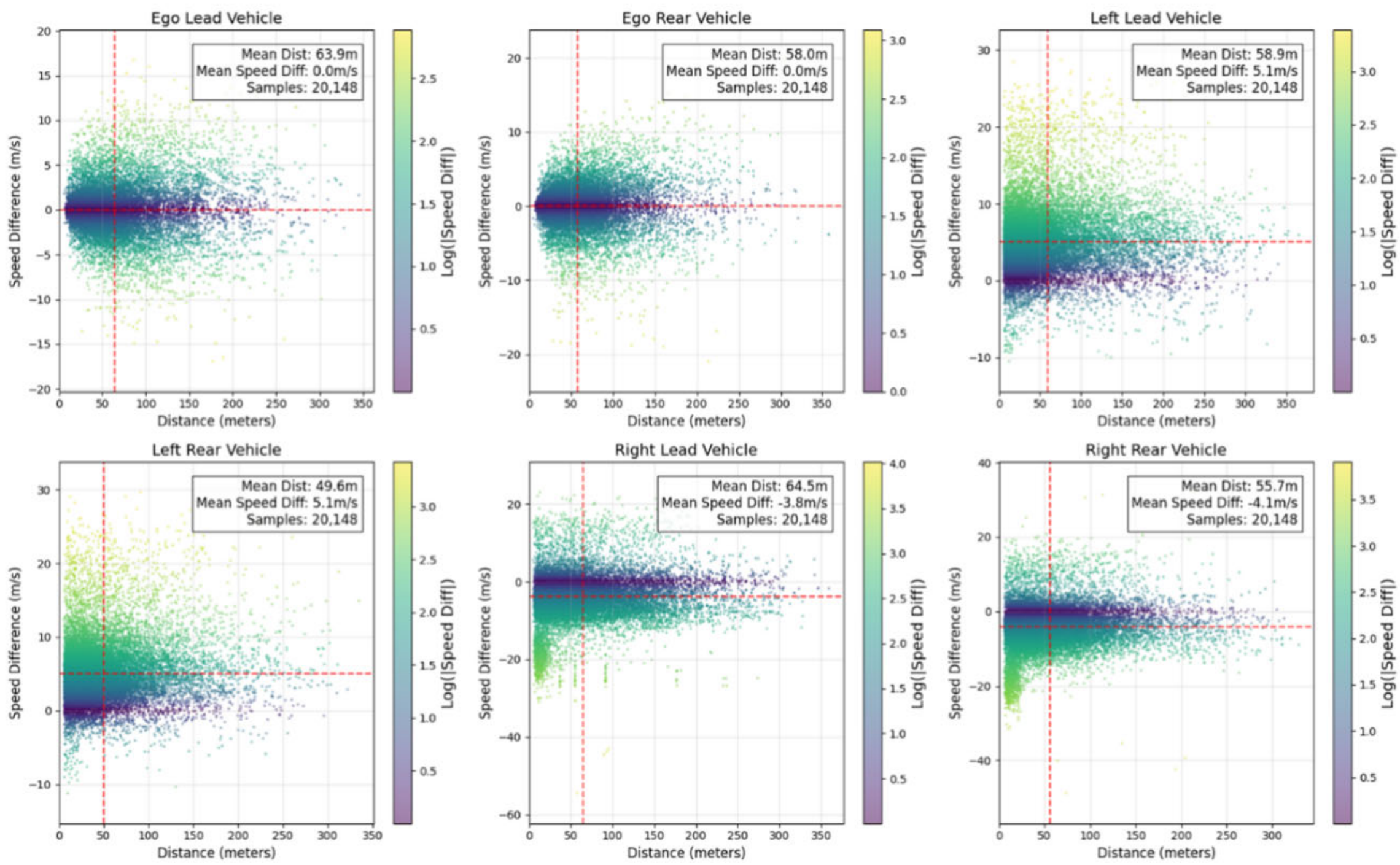

Figure 8. Distance-speed difference distribution (based on the exiD dataset).

Figures 7 and 8 illustrate the joint distribution of lateral distances and relative speed differences between the ego vehicle and its six surrounding neighbors at the moment of a lane change. Most data points cluster around a near-zero speed difference, while a smaller subset falls into the "large speed difference / short distance" region, indicating potential risk-prone scenarios. To transform this two-dimensional distribution into a learnable variable, we introduce the Closing Gap Time (CGT), defined for each neighbor $i$ as:

$$\mathrm{CGT}_i = \frac{d_i}{|\Delta v_i| + \varepsilon} \tag{16}$$

where $d_i$ is the lateral distance, $\Delta v_i$ the relative speed between the ego and neighbor vehicles, and $\varepsilon$ a small stabilizing constant. CGT directly reflects the time required for the current speed difference to close the gap, thereby serving as an indicator of lane-cha0nge safety margins (with smaller values indicating higher risk).

4. Longitudinal safety metrics. We further incorporate internal safety indicators such as *Distance Headway (DHW)*, *Time Headway (THW)*, and *Time-to-Collision (TTC)*. For each metric, we extract the minimum value observed within a historical window, using it as a hard or soft safety constraint in the loss function.

5. Driving behavior semantics. To capture driver intent and style, we derive features including vehicle type, frequency of past lane changes, roadway speed limits, instantaneous speed, speed ratio, and acceleration ratio. These descriptors encode behavioral tendencies that influence lane-change decisions, such as aggressiveness or cautiousness.

In addition to these general modules, the exiD dataset provides extended annotations based on OpenDRIVE roadway topology. This enables the inclusion of ramp-specific features, such as distance to ramp entry/exit, estimated time-to-arrival, and multi-horizon indicators (5 s, 15 s, and 30 s). These additional attributes significantly enhance the model's ability to anticipate lane-change intentions in complex merging and diverging environments.

## 4. EXPERIMENTS

### 4.1. Experimental setup

To systematically evaluate how historical observation length and prediction horizon affect lane-change intention recognition, we designed a series of comparative experiments on the highD and exiD real-world trajectory datasets. To ensure strict independence between training and testing samples, and to avoid trajectory leakage or source bias, both datasets were partitioned by locationId: for highD, data recorded at locations 0, 1, 2, 3 were used for training, while data from locations 4 and 5 were reserved for testing; for exiD, training samples were drawn from locations 0, 1, 2, and 3, and testing samples came from locations 4, 5, and 6.

Before training, all models underwent hyperparameter tuning via Bayesian Optimization combined with 5-fold cross-validation. This setup enabled efficient exploration of the parameter space within limited computational budgets.

Among traditional machine learning methods, we conducted a preliminary comparison between LightGBM and XGBoost. The results indicated that while XGBoost achieved only marginal improvements in accuracy ( ≤ 0.05 percentage points on highD and ≤ 0.4 on exiD), it incurred substantially higher computational costs—training time on exiD was approximately 6× that of LightGBM. Balancing accuracy with efficiency, we therefore adopted LightGBM as the representative traditional model in subsequent experiments.

To comprehensively analyze the effect of observation length on predictive performance, we evaluated multiple history window settings across prediction horizons of 1 s, 2 s, and 3 s. The history window length ranged from 1–5 s for highD and 2–7 s for exiD, ensuring coverage of plausible optimal intervals. For each prediction horizon T, we exhaustively tested all candidate history windows W, recording both training and generalization accuracy, and subsequently identified the optimal history window for each T.

### 4.2. Performance across historical windows, prediction horizons, and models

#### 4.2.1. LightGBM results

On the highD dataset, the best-performing history window was W = 1s. At prediction horizons of T = 1s, 2s, and 3s, the model achieved generalization accuracies of 0.9981, 0.9952, and 0.9906, respectively, with corresponding Macro F1-scores of 0.9359, 0.9126 and 0.8761. As shown in Table 1, the No-LC class consistently achieved F1-scores close to 1.0, while the Left-LC and Right-LC classes showed declines at the 2-second horizon and 3-second horizon.

In contrast, the exiD dataset yielded its best results with a longer history window, W = 6s. At T = 1s, 2s, and 3s, the generalization accuracies were 0.9607, 0.8950, and 0.8117, with Macro F1-scores of 0.9004, 0.8343, and 0.7819. Both accuracy and F1 dropped steadily as the prediction horizon lengthened, with a marked degradation observed at $T = 3\ s$.

#### 4.2.2. Two-layer stacked LSTM results

For the highD dataset, the optimal history window was W = 2s. Under these settings, the generalization accuracies were 0.9970, 0.9934, and 0.9875 for the three horizons, with Macro F1-scores of 0.9162, 0.8847, and 0.8430, respectively (Table 2). Similar to LightGBM, the No-LC class maintained near-perfect F1, while Left-LC and Right-LC recognition performance declined at T = 2s and T = 3s.

On the exiD dataset, the best history windows were W = 4s for T = 1s and 2s, and W = 5s for T = 3s. The corresponding generalization accuracies were 0.8705, 0.7742, and 0.7058, with Macro F1-scores of 0.7702, 0.7233, and 0.6875. As summarized in Table 2, performance consistently decreased with longer prediction horizons, and the declines were particularly pronounced for the Left-LC and Right-LC classes.

Table 1. Optimal performance of LightGBM across different prediction horizons.

| Dataset | Prediction Horizon | Overall Accuracy | Macro F1 | NO-LC F1 | Left-LC F1 | Right-LC F1 |
|---|---|---|---|---|---|---|
| highD | 1s | 0.9981 | 0.9359 | 0.9990 | 0.8999 | 0.9087 |
| highD | 2s | 0.9952 | 0.9126 | 0.9975 | 0.8588 | 0.8815 |
| highD | 3s | 0.9906 | 0.8761 | 0.9951 | 0.8017 | 0.8315 |
| exiD | 1s | 0.9607 | 0.9004 | 0.9895 | 0.8512 | 0.8606 |
| exiD | 2s | 0.895 | 0.8343 | 0.9755 | 0.7465 | 0.781 |
| exiD | 3s | 0.8117 | 0.7819 | 0.9284 | 0.7025 | 0.7148 |

Table 2. Optimal performance of the two-layer stacked LSTM across different prediction horizons.

| Dataset | Prediction Horizon | Overall Accuracy | Macro F1 | NO-LC F1 | Left-LC F1 | Right-LC F1 |
|---|---|---|---|---|---|---|
| highD | 1s | 0.9970 | 0.9162 | 0.9981 | 0.8715 | 0.8790 |
| highD | 2s | 0.9934 | 0.8847 | 0.9964 | 0.8024 | 0.8553 |
| highD | 3s | 0.9875 | 0.8430 | 0.9930 | 0.7399 | 0.7961 |
| exiD | 1s | 0.8704 | 0.7702 | 0.9284 | 0.6717 | 0.7104 |
| exiD | 2s | 0.7742 | 0.7233 | 0.8657 | 0.6462 | 0.6579 |
| exiD | 3s | 0.7058 | 0.6875 | 0.8020 | 0.6264 | 0.6341 |

### 4.3. Evaluation and analysis

From the overall results, two factors stand out as the primary determinants of model performance: the prediction horizon ($T$) and the complexity of the traffic scene.

#### 4.3.1. Effect of prediction horizon

In the highD dataset, both LightGBM and the two-layer stacked LSTM achieved accuracy consistently at or above 99%. However, this high accuracy largely reflects the severe class imbalance: even after applying imbalance-handling strategies, the disparity among classes remained substantial, with non-lane-change samples dominating at an approximate ratio of 29:1:1 compared to left and right lane changes. Under such conditions, a model biased toward predicting "no lane change" can still achieve superficially high accuracy. Macro F1 provides a more reliable assessment: as the prediction horizon increased, Macro F1 declined from 0.9359 at T = 1s to 0.8761 at T = 3s for LightGBM, reflecting a clear reduction in the model's ability to recognize lane-change maneuvers.

In the exiD dataset, where the class distribution was more balanced (ratios ranging from roughly 8:1:1 to 2:1:1), the downward trends in accuracy and Macro F1 were more closely aligned. As the horizon lengthened from 1 s to 3 s, accuracy fell from 0.9607 to 0.8117, and Macro F1 dropped from 0.9004 to 0.7819. These results highlight the increasing difficulty of making reliable predictions further in advance, especially under complex traffic conditions.

#### 4.3.2. Comparison of models

When comparing the two models, LightGBM and the two-layer stacked LSTM each show distinct strengths and weaknesses depending on the dataset and prediction horizon.

In the highD dataset, where the driving environment is relatively simple and the data distribution is highly imbalanced, both models achieved near-perfect accuracies. While their accuracies were almost indistinguishable, LightGBM delivered stronger macro F1-scores at all horizons (e.g., 0.9359 vs. 0.9162 at T=1s), suggesting that tree-based methods handle majority-class dominance with greater stability. However, the performance gap between the two-layer stacked LSTM and LightGBM remained narrow in this setting. This outcome suggests that the temporal modeling capability of the two-layer stacked LSTM, which is its principal strength in capturing sequential dependencies, provides limited incremental benefit when driving patterns are straightforward, leading to performance that is largely comparable to that of tree-based approaches. Given that the LSTM model in this study adopts a concise two-layer configuration optimized for computational efficiency rather than architectural depth, its temporal modeling capacity may not be fully exploited under the current setup.

On the exiD dataset, which features a more complex traffic environment and a more balanced class distribution, the two-layer stacked LSTM's sequential modeling ability did not translate into a clear edge at short horizons. At T=1s, its generalization accuracy (0.8705) and macro F1 (0.7702) were both below LightGBM's (0.9607 and 0.9004). As the prediction window extended, the two-layer stacked LSTM's performance deteriorated more rapidly, with macro F1 dropping to 0.6875 at T=3s, whereas LightGBM, though also declining, maintained a relatively higher score of 0.7819. This highlights LightGBM's greater robustness in more diverse and challenging driving scenarios.

Overall, LightGBM consistently demonstrates stronger stability and generalization across settings, while the two-layer stacked LSTM struggles with longer horizons and complex environments. These findings suggest that under the current data conditions, traditional machine learning models—leveraging well-designed features and simpler architectures—can still outperform deep learning approaches.

#### 4.3.3. Class-level performance

At the class level, the No-LC category consistently achieved markedly stronger results relative to the other classes, reflecting strong and stable recognition of the majority class. By contrast, the F1-scores for Left-LC and Right-LC declined with longer prediction horizons, particularly on exiD at T = 3s. This suggests that the erosion of performance in minority classes is the main driver of the overall decrease in Macro F1.

#### 4.3.4. Summary

Based on the experimental results of this study, we can draw three conclusions regarding vehicle lane change intention prediction on multi-scenario highways:

**1.Model Selection**

From a model selection perspective, traditional machine learning methods still demonstrate strong competitiveness. Although deep learning models theoretically have an advantage in time series modeling, this advantage was not fully realized under the conditions of imbalanced training data set and comprehensive feature engineering. Notably, LightGBM showed greater robustness in handling imbalanced classes and cross-scenario generalization, indicating that feature-driven, concise structures still hold high practical value for real-world deployment.

**2.Traffic Scenario**

Differences From the perspective of traffic scenario differences, straight road sections and ramp areas exhibit significant disparities in sample distribution, interaction complexity, and prediction difficulty. The model's performance on the highD dataset was significantly better than its performance on the exiD dataset, with higher F1 scores. This suggests that the difficulty of lane change prediction on straight highway sections is lower than in highway ramp merge/diverge areas. This phenomenon indicates that future lane change prediction research should fully account for scenario differences and avoid directly generalizing results from a single scenario to more complex environments.

**3.Task Definition**

Finally, from the task definition perspective, extending the lane change behavior to a three-class classification (stay in lane, change left, change right) offers a significant advantage in identifying directional differences. However, the deficiency in predicting minority classes remains a major bottleneck for performance degradation, which becomes more pronounced with longer prediction horizons. This finding suggests that a high-precision minority class modeling capability is a critical breakthrough for improving the overall system reliability.

## 5. CONCLUSION AND FUTURE WORK

This study addresses the problem of highway lane-change intention prediction using naturalistic trajectory data from the highD and exiD datasets. Lane-change behavior is formulated as a three-class task (left, right, none), and a physics-informed framework is proposed that embeds roadway dynamics and traffic-safety principles into both feature design and model training. Experiments across 1–3 s horizons show that LightGBM consistently outperforms the two-layer stacked LSTM in efficiency and cross-scenario robustness, making it better suited for real-time applications. While the two-layer stacked LSTM can capture temporal dependencies, its performance suffers from class imbalance and data sparsity, limiting generalization. A common trend is that prediction accuracy declines with longer horizons, especially for minority classes, and performance further degrades in complex environments such as ramps, underscoring the challenge of mid- to long-term intention prediction under high interaction density.

Future work will proceed in three directions.

1. **LSTM architecture enhancement.** Future work will investigate deeper or attention-augmented recurrent structures built upon the current two-layer configuration, aiming to better capture long-term temporal dependencies and improve robustness under complex or long-horizon conditions.
2. **Hybrid modeling.** Combine fast tree-based models for confident cases with temporal models for ambiguous sequences to balance efficiency and accuracy.
3. **Stronger physics integration.** Incorporate richer monotonic constraints, differentiable safety margins, and feasibility priors to strengthen physical plausibility.

4. **Minority-class accuracy improvement.** Enhance recognition of lane changes through tailored loss functions, class-conditional calibration, augmentation, and horizon-adaptive strategies.

Overall, the proposed physics-informed framework provides a practical and interpretable path to reliable multi-scenario lane-change intention prediction, and lays the groundwork for safer decision making in intelligent connected vehicles.